\documentclass[review]{elsarticle}
\graphicspath{ {./figures/} }
\usepackage{hyperref}
\usepackage{float}
\usepackage{verbatim} 
\usepackage{apalike}
\restylefloat{figure}
\restylefloat{table}
\usepackage{algorithm}
\usepackage{algorithmic}
\usepackage{xcolor}
\usepackage{tabularx} 
\usepackage{amsmath}

\usepackage{listings}
\definecolor{codegreen}{rgb}{0,0.6,0}
\definecolor{codegray}{rgb}{0.5,0.5,0.5}
\definecolor{codepurple}{rgb}{0.58,0,0.82}
\definecolor{backcolour}{rgb}{0.95,0.95,0.92}
\lstdefinestyle{mystyle}{
    backgroundcolor=\color{backcolour},   
    commentstyle=\color{codegreen},
    keywordstyle=\color{magenta},
    numberstyle=\tiny\color{codegray},
    stringstyle=\color{codepurple},
    basicstyle=\ttfamily\footnotesize,
    breakatwhitespace=false,         
    breaklines=true,                 
    captionpos=b,                    
    keepspaces=true,                 
    numbers=left,                    
    numbersep=5pt,                  
    showspaces=false,                
    showstringspaces=false,
    showtabs=false,                  
    tabsize=2
}
\journal{Preprint for arxiv}

\biboptions{authoryear}

\begin{document}
\begin{frontmatter}

\begin{titlepage}
\begin{center}
\vspace*{1cm}



\textbf{ \large Evaluating LLM - Generated Multimodal Diagnosis from Medical Images and Symptom Analysis }

\vspace{1.5cm}

Dimitrios P. Panagoulias$^{a}$ (panagoulias\_d@unipi.gr), Maria Virvou$^{a}$ (mvirvou@unipi.gr), George A. Tsihrintzis$^{a}$ (geoatsi@unipi.gr) \\

\hspace{10pt}

\begin{flushleft}
\small  
$^a$ Department of Informatics, University of Piraeus 185 34, Greece \\

\vspace{1cm}
\textbf{Corresponding Author:} \\
George A. Tsihrintzis \\
Department of Informatics, University of Piraeus 185 34, Greece \\
Tel: (+30) 697 2882168 \\
Email: geoatsi@unipi.gr

\end{flushleft}        
\end{center}
\end{titlepage}
\title{Evaluating LLM - Generated Multimodal Diagnosis from Medical Images and Symptom Analysis}

\author[label1]{Dimitrios P. Panagoulias}
\ead{panagoulias\_d@unipi.gr}

\author[label1]{Maria Virvou}
\ead{mvirvou@unipi.gr}

\author[label1, cor1]{George A. Tsihrintzis}
\ead{geoatsi@unipi.gr}

\cortext[cor1]{Corresponding author.}
\address[label1]{Department of Informatics, University of Piraeus, Piraeus 185 34, Greece}

\begin{abstract}
Large language models (LLMs) constitute a breakthrough state-of-the-art Artificial Intelligence technology which is rapidly evolving and promises to aid in medical diagnosis. However, the correctness and the accuracy of their returns has not yet been properly evaluated. In this work, we propose an LLM evaluation paradigm that incorporates two independent steps of a novel methodology, namely (1) multimodal LLM evaluation via structured interactions and (2) follow-up, domain-specific analysis based on data extracted via the previous interactions. Using this paradigm, (1) we evaluate the correctness and accuracy of LLM-generated medical diagnosis with publicly available multimodal multiple-choice questions(MCQs) in the domain of Pathology and (2) proceed to a systemic and comprehensive analysis of extracted results. We used GPT-4-Vision-Preview as the LLM to respond to complex, medical questions consisting of both images and text, and we explored a wide range of diseases, conditions, chemical compounds, and related entity types that are included in the vast knowledge domain of Pathology. GPT-4-Vision-Preview performed quite well, scoring approximately 84\% of correct diagnoses. Next, we further analyzed the findings of our work, following an analytical approach which included Image Metadata Analysis, Named Entity Recognition and Knowledge Graphs. Weaknesses of GPT-4-Vision-Preview were revealed on specific knowledge paths, leading to a further understanding of its shortcomings in specific areas. Our methodology and findings are not limited to the use of GPT-4-Vision-Preview, but a similar approach can be followed to evaluate the usefulness and accuracy of other LLMs and, thus, improve their use with further optimization.
\end{abstract}

\begin{keyword}
AI-empowered software engineering \sep Multimodal LLMs \sep GPT-4-Vision \sep Telehealth \sep Medical diagnosis
\end{keyword}
\end{frontmatter}
\begin{center}
  \includegraphics[width=1.0\textwidth]{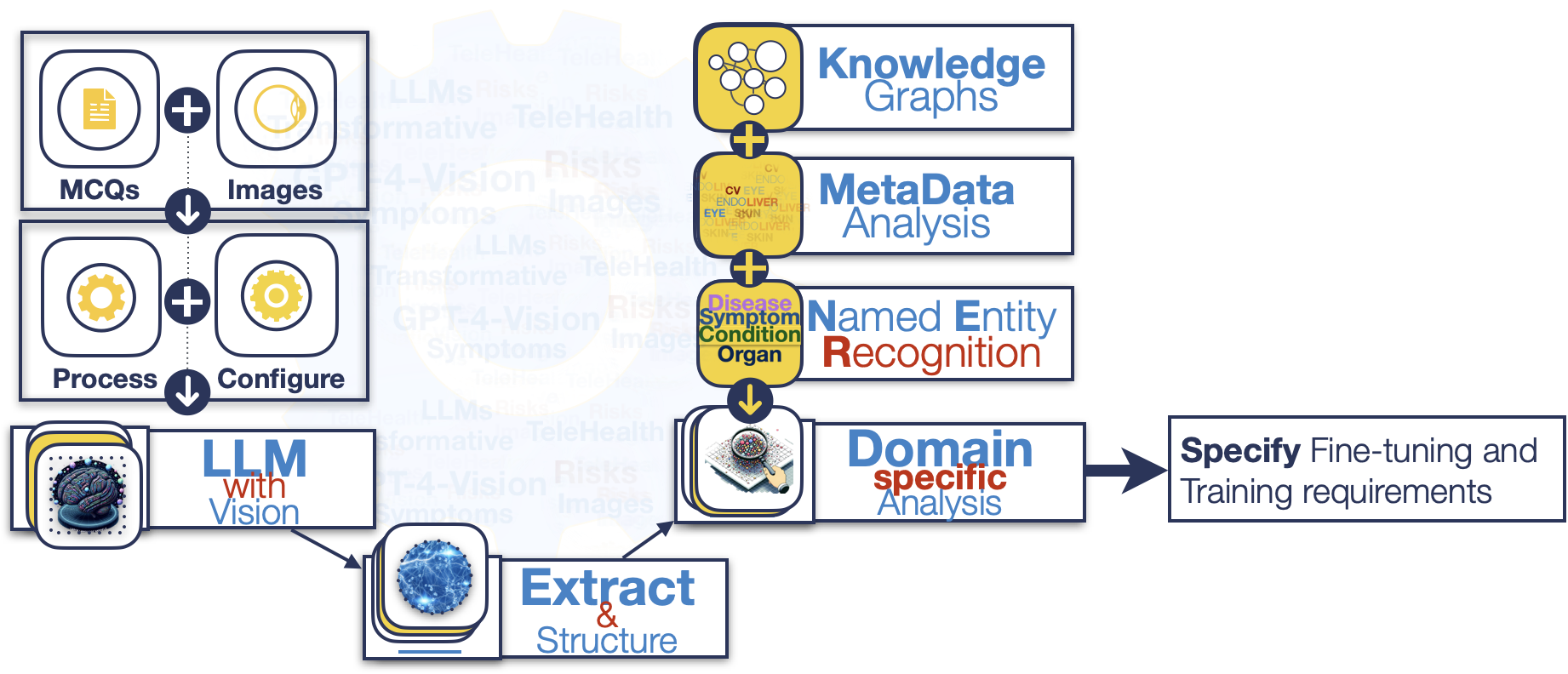}
\end{center}

\section{Introduction}
\label{introduction}

Streamlined digital technologies, empowered by Artificial Intelligence (AI) and Big Data Analytics (BDA), are increasingly demonstrating their potential to enhance  symptom detection, provide accurate and fast medical diagnosis (e.g.  \cite{stokes2022use}) and predict treatment outcomes (e.g. \cite{alanazi2017critical}). Indeed, AI acts as a pivotal force in aligning user (patient or medical personnel) needs with the development of tools which can efficiently and reliably analyze health-related data and biosignals and either lead to entirely new medical approaches or enhance existing ones. This, in turn, improves the quality of life, extends lifespans, and positively impacts both Quality-Adjusted Life Years (QALYs) and Disability-Adjusted Life Years (DALYs) \citep{carrandi2023systematic}.

However, many challenges emerge with all AI-empowered systems, in terms of relevance and potential misuse. Such challenges include issues of validation/evaluation that stem from confirmation bias or biases of those responsible for the system's validation, evaluation and management. Thus, a need arises for transparency, rigorous evaluation and continuous improvement. Given the high complexity of medical recommendations, associated interventions carry significant risks and potential threats to life and well-being. Clearly, ensuring patients' health and data safety necessitates frameworks that clarify and interpret all underlying technologies as AI-influenced decisions could have very significant consequences \citep{rasheed2022explainable, panagoulias2023tailored}. 

In this paper, we propose an LLM evaluation paradigm that incorporates two independent steps of a novel methodology, namely (1) multimodal LLM evaluation via structured interactions and (2) follow-up, domain-specific analysis based on data extracted via the previous interactions. Next, using this paradigm, we evaluate the correctness and accuracy of LLM-generated medical diagnosis with publicly-available multimodal multiple-choice questions in the domain of Pathology, followed by a systemic and comprehensive analysis of extracted results \citep{utah}.

We use GPT-4-Vision-Preview \citep{gpt44} (training data Up to Apr 2023), a state-of-the-art LLM based on the GPT transformer, as the LLM to respond to complex, medical questions consisting of both images and text. Our work is focused on Pathology which implies that we are not looking at only one specific disease, condition, medical domain or specialty. Instead, we are exploring a wide range of diseases, conditions, chemical compounds and related entity types that are included in the vast knowledge domain of Pathology. GPT-4 Vision performed quite well in our evaluation, scoring approximately 84\% of correct diagnoses. This high level of accurate diagnosis across diverse areas, which typically requires a variety of deep learning and machine learning tools, incentivized us to further analyze the findings of our work following an analytical approach. Our analytical approach included Image Metadata Analysis (IMA), Named Entity Recognition (NER) and Knowledge Graphs (KG), which revealed weaknesses of GPT-4-Vision-Preview on specific knowledge paths and allowed a further understanding of the employed LLM's shortcomings in specific areas. Our methodology and findings are not limited to the use of GPT-4-Vision-Preview, but a similar approach can be followed to evaluate the usefulness and accuracy of other LLMs and, thus, improve their use with further optimization.

More specifically, the paper is structured as follows: Section~2 focuses on reviewing relevant previous studies and related published work. In Section~3, the evaluation methodology is presented and consists of two independent parts. The first part is related to specific interactions with a multimodal LLM for defining competency using structured multiple-choice questions. The second part is related to the analysis of those interactions to extract useful information for enhancing the LLM knowledge base. In Section~4, we present the results from application of the proposed methodology in Pathology utilizing GPT-4-Vision-Preview and multiple-choice questions and images. Lastly in Section~5, we summarize the findings of our paper and discuss future related endeavours.

\section{Related work}

In this section, we establish a comprehensive background pertinent to the methodology used in this work and the various aspects of relevant technologies. We also briefly describe a recent previous work of ours, in which we evaluated GPT-4's ability in General Pathology Multiple-Choice Quizzies (MCQs) from only provided textual descriptors of symptoms \citep{panagoulias2024augmenting}. 

\subsection{Evaluation of medical students}
\label{comGap}
While using  new technologies for automating the provision of medical services is very promising, the tolerance for mistakes is extremely low and the risk remains high of misdiagnosis, privacy breaches, and ethical dilemmas such as bias in AI algorithms. 

Currently, for the comprehensive evaluation of medical students, various assessment methods are used alongside Objective Structured Clinical Examinations (OSCEs) \citep{majumder2019evaluative} and multiple-choice quizzes \citep{vanderbilt2013assessment, norcini2007assessment, ferris2015assessment}. Each of these methods has its own focus, ranging from theoretical knowledge to practical skills and professional behaviors. 

More specifically, a list of common assessment procedures followed in medical education is presented next: 
\begin{enumerate}
    \item Objective Structured Clinical Examinations (OSCEs): They assess clinical skills, communication, and procedures in a structured and practical setting.
    \item Multiple-choice Quizzes (MCQs): They test theoretical knowledge, understanding of concepts, and problem-solving abilities.
    \item Short Answer Questions (SAQs): They valuate the ability to provide concise, precise answers to focused questions.
    \item Essay Questions: They assess depth of understanding, ability to synthesize information, and critical thinking.
    \item Oral Examinations/Vivas: They test understanding and knowledge through direct questioning and are often used in higher level assessments like specialty/doctoral exams.
    \item Practical or Lab-based Assessments: They evaluate skills in laboratory techniques, clinical procedures, or other practical tasks.
    \item Direct Observation of Procedural Skills (DOPS): They assess specific procedural skills in a clinical setting with direct observation by an examiner.
    \item Case-based Discussions (CBDs): Evaluate clinical reasoning, decision-making, and professional judgment using real or simulated patient cases.
    \item Portfolio Assessments: They compile and review a collection of a student's work over time to assess his/her progress and competence.
    \item 360-Degree Feedback/Peer Assessment: They gather feedback from a wide range of observers, including peers, patients, and other healthcare professionals, to assess communication, teamwork, and professional behavior.
    \item Self-Assessment and Reflection: They encourage students to reflect on their own learning and performance.
    \item Standardized Patient Encounters: They use actors trained to portray real patient cases for assessment of clinical and communication skills.
    \item Logbooks or Procedure Lists: They track the number and type of clinical procedures or experiences that a student has completed.
    \item Audit Projects or Quality Improvement Projects: They evaluate the ability to conduct research, analyze data, and understand the principles of improving clinical practice.
\end{enumerate}
The evaluation and training of medical students involve a complex process, indicating that the MCQs utilized in this study represent only a small portion of the evaluation processes necessary for certifying a doctor.

\subsection{Vision transformers}

Vision Transformers (ViTs) \citep{zhai2022scaling} constitute a type of artificial intelligence model that has revolutionized the field of computer vision, i.e. the AI discipline focused on enabling machines to interpret and understand visual information from the world \citep{khan2022transformers}. ViTs are, in fact, a variation of the Transformer model \citep{vaswani2017attention, devlin2018bert}, which was originally developed for NLP tasks. 

During training \citep{atito2021sit}, ViTs learn to identify features and patterns within the patches, leveraging the self-attention mechanism to understand the context and relationships between different parts of an image \citep{yuan2021tokens, zhai2022scaling}. 
For inference, the ViT model processes a new image through the same steps as used in its training and the transformed class token is used to make predictions or classifications. The key advantage of ViTs over traditional Convolutiona Neural Networks (CNNs) is their ability to capture global dependencies across the entire image, thanks to their self-attention mechanism. This allows for a more holistic understanding of the image context and leads to significant improvements in various computer vision applications.

Similarly to the traditional transformer, training or fine-tuning \citep{sandler2022fine} a ViT can be done (1) either by providing pair-wise inputs of images and their corresponding annotations or (2) by using self-supervised learning techniques. In the former approach, the model learns from labeled datasets where each image is paired with relevant information, such as classification labels or bounding boxes for object detection. In the latter case, the model leverages unlabeled data, autonomously extracting patterns and features through methods like contrastive learning or predictive coding, which help in understanding the visual context without explicit human-provided labels. The effectiveness of ViTs hinges not just on the architecture of the model itself, but significantly on how the training data is curated and presented. Indeed, cross-modal training plays a pivotal role in bridging the gap between different types of data.

The remarkable performance of the Transformer architecture in NLP has recently also triggered broad interest in their application in Computer Vision \citep{azad2023advances}. Among other merits, Transformers are witnessed as capable of learning long-range dependencies and spatial correlations, which is a clear advantage over the commonly-employed CNNs \citep{li2014medical}.

\subsection{Named Entity Recognition and Knowledge Graphs}
Named Entity Recognition (NER) \citep{nadeau2007survey, hong2020sciner} is a sub-field of NLP and Information Retrieval. The primary goal of NER is to identify and categorize key information in text into predefined categories or ``entities" such as the names of persons, organizations, locations, expressions of times, quantities, monetary values, percentages, etc. NER systems are designed to recognize entities within a body of text. This involves not only detecting the entity, but also categorizing it into a predefined class. 

A Knowledge Graph (KG) \citep{ji2021survey, abu2021domain} is a structured way of representing data in a graph format, where entities (nodes) are connected by relationships (edges). This graphical representation allows for the efficient organization, management, and retrieval of complex information. Network density is calculated as the number of existing edges in the network divided by the number of all possible edges. The exact formula depends of whether a KG is directed or undirected.

For undirected KGs (which are also used in this work), network density is calculated via the following formula: 

\begin{center}
\( D = \frac{2E}{N(N - 1)} \),
\end{center}
\begin{center}
where
\end{center}
\begin{center}
D  \text{ is the density of the graph,}\\
E  \text{ is the number of edges in the graph,}\\
N  \text{ is the number of nodes in the graph.} \\
\end{center}

On the other hand, for a directed graph, each possible edge is counted twice, i.e. separately for each direction. Thus, the factor 2 in the previous formula needs to be removed and the network density is calculated as
: 
\begin{center}
\( D = \frac{E}{N(N - 1)} \).
\end{center}

NER is often a crucial first step in the creation and enrichment of KGs. By identifying entities like names of people, organizations, locations, etc., NER systems can feed this structured information into a KG. Each identified entity can become a node in the graph and the context or relationships between these entities can form the edges. 

\subsection{Evaluation of GPT-4.0 in a General Pathology questionnaire}
In our recent work\citep{panagoulias2024augmenting}, we evaluated the level of proficiency of GPT-4 in answering General Pathology questions in the form of MCQs. The GPT-4 model achieved a total precision of 91.37\%, answering correctly 180 out of 197 questions from the subdomains of Arterosclerosis and Thrombosis, Cellular Injury, Embryology, and Nutrition. The MCQ questions were retrieved from `The Internet Pathology Laboratory for Medical Education', an~esteemed resource hosted by the University of Utah's Eccles Health Sciences Library~\citep{utah}.

\section{Materials and Methods}
In this section, we define and detail the main components of our evaluation methodology and the proposed analytical services to define fine-tuning or retraining requirements based on specific knowledge paths. Our methodology combines elegant approaches for targeted and complex data analysis. These include data pre-processing, prompt engineering, and data extraction and refinement. Our methods encompass IMA, scientific NER, knowledge extraction, and investigation through KGs.

Our proposed methodology is divided into two distinct parts. The first part involves the `Multimodal Large Language Model (LLM) Evaluation', where we focus on the aspects of interaction with the model and the methods used for data extraction. This stage is crucial for understanding how the model processes and responds to various inputs. In the second part, we conduct a `Domain specific Analysis'. This phase is dedicated to identifying the specific needs and requirements for further fine-tuning or re-training of the model to better suit domain-specific applications. This approach ensures a comprehensive evaluation and enhancement of the model's performance in targeted areas.

\subsection{Multimodal LLM evaluation}
\begin{figure}[H]
    \centering
    \includegraphics[width=0.8\linewidth]{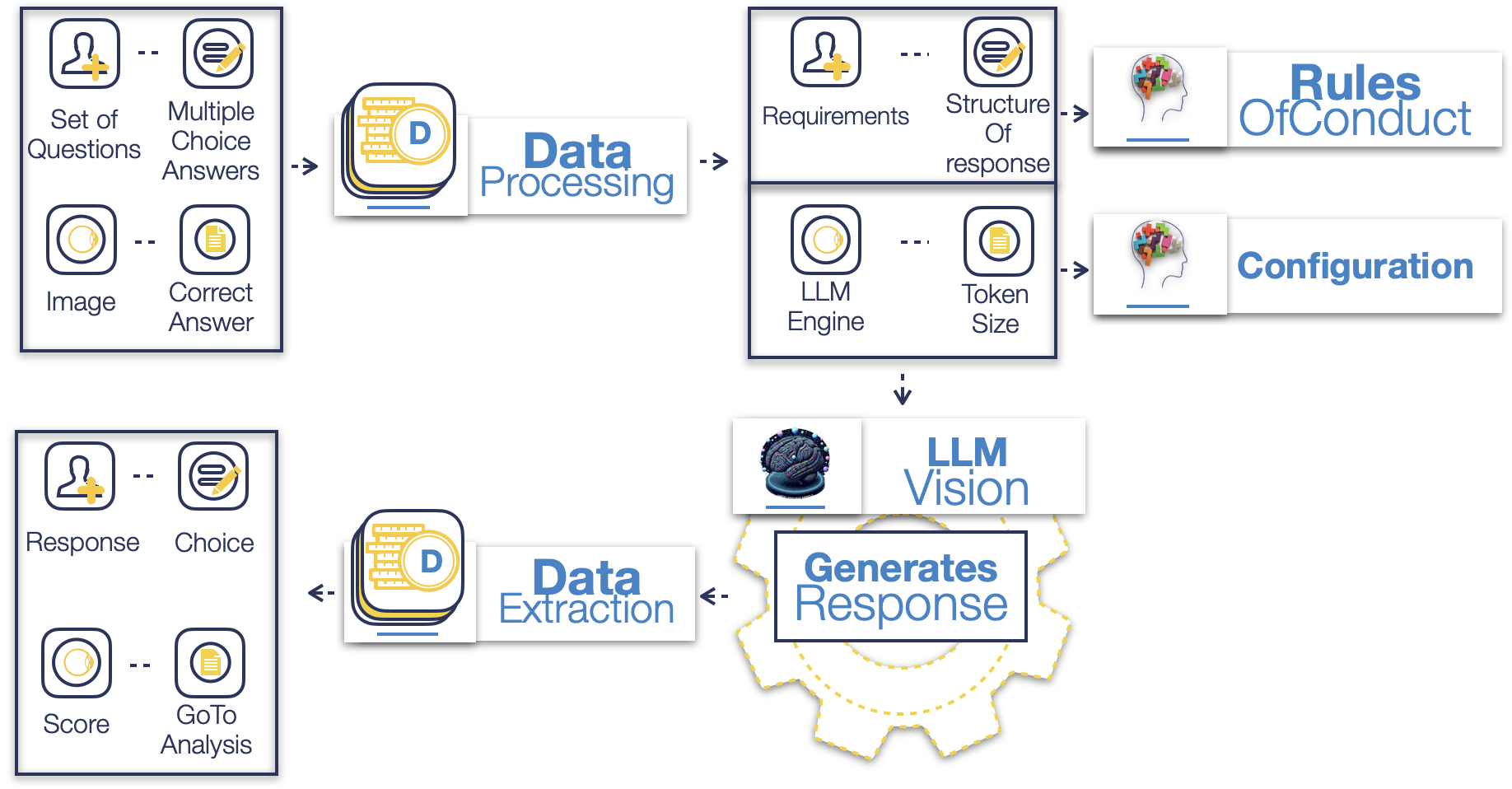}
    \caption{Multimodal LLM evaluation}
    \label{fig:procOut}
\end{figure}

As can be seen in figure~\ref{fig:procOut}, the first step of the Multimodal LLM evaluation is data processing. Here, we select the set of MCQs and answers, as per specific domain prerequisites. The set of questions are associated with the image file, which is related to the question, alongside the correct answer. The processed data are provided as external files and then engineered in the required format for interaction with the selected LLM.

In the second step, the \textit{Rules Of Conduct} are defined, which are the essential parts of the prompt enginneering, which requests the specific format of analysis and response from the chosen LLM. The \textit{Configuration} part defines the chosen model, highlighted in figure~\ref{fig:procOut} as \textit{LLM Engine} where also the \textit{Token Size} of the response is set. The \textit{LLM} model with \textit{Vision} is provided in the format (image, question, rules of conduct and configuration settings) and a structured response is generated.

This part of our methodology is designed to enhance the evaluation process of a LLM  with vision, focusing on its multimodal capabilities. It facilitates the creation of structured tests and quizzes that integrate both images and related questions, presenting a holistic challenge to the model's interpretative abilities. In each instance, the model is required to analyze the visual content alongside the accompanying question to select the most appropriate answer from multiple choices. This approach ensures a thorough assessment of the model's proficiency in interpreting and correlating different types of data.

Following the completion of the test, a detailed score is provided. This score reflects the model's performance, highlighting the correctly identified answers, while offering explanations for any incorrect responses. This feature is particularly valuable as it sheds light on areas where the model may need further refinement and, thereby, guides towards future improvements. 

\subsection{Domain-specific analysis}
The data extracted previously are used in the domain-specific analysis. This part of our evaluation methodology, includes 3 main steps that ensure the definition of requirements of fine-tuning and training to make the  multimodal LLM excel where it scored lower. 

As can be seen in figure~\ref{fig:procOut}, in the first step IMA takes place and evaluates the domains where the multimodal responded correctly and those where it responded incorrectly. 

In the second step, we apply NER on both the correct and the incorrect responses. Specifically, NER is applied on the correct responses and on the solution explanation of the incorrect responses. Here the exported report includes entities grouped based on examined response or explanation, correspondingly.

Lastly, in the third step, the extracted entities are plotted into KGs. The report in this step is based on the evaluation of connection density and the information spread potential. 

From all steps, the specifications for fine-tuning and training requirements are defined and provide improvement steps for the multimodal LLM in the examined domain.

\begin{figure}[H]
    \centering
    \includegraphics[width=1.1\linewidth]{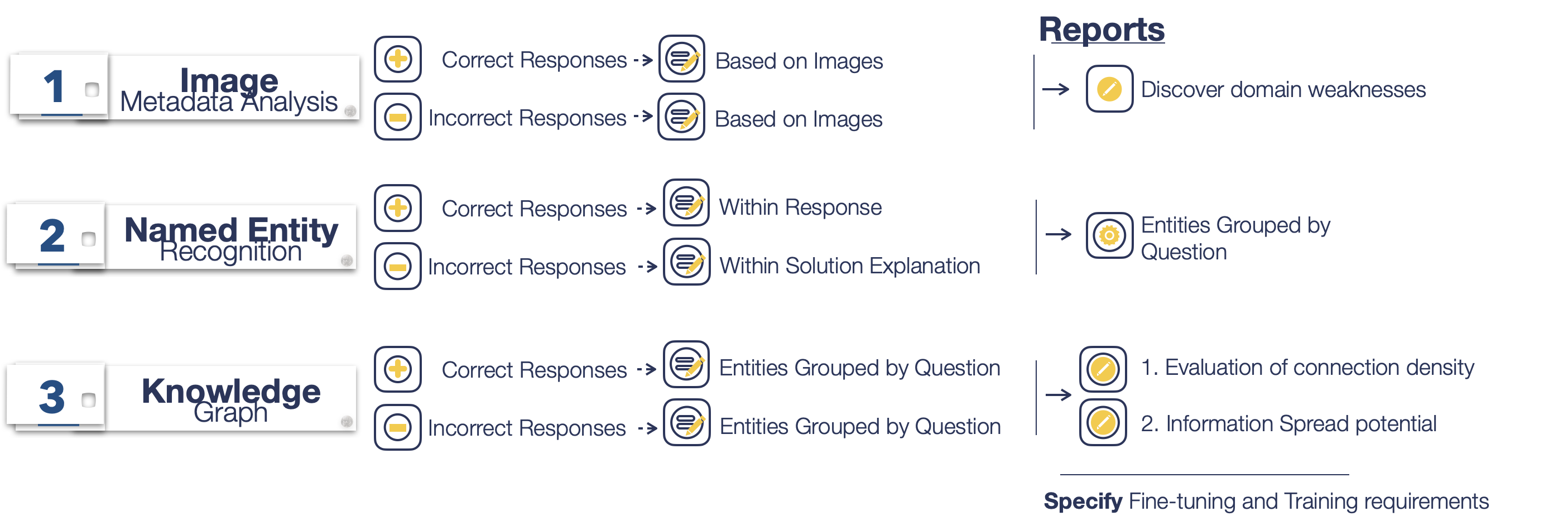}
    \caption{Domain specific Analysis}
    \label{fig:analProc}
\end{figure}

\section{Results}
\subsection{Pathology Quiz with Images for Multimodal Evaluation of LLMs}
Using the previously-described methodology, a Pathology quiz with images was used and our proposed methodology is applied. The results are reported here. 

To facilitate the reader, we provide the previously-described methodology, as packaged python libraries, available as open source with associated materials used for this analysis. 

The MCQ, that can be found here \citep{utah}, is constructed to test medical students in a variety of sub-domains. Each sub-domain includes 10 questions with symptomatology, a related medical image and multiple-choice answers of which only one is correct. It should be mentioned that at the moment of writing ChatGPT, which is OpenAI's commercially available chatbot powered by its latest models, does not accept medical images for analysis. Thus, we developed a software module that interacts with hosted LLMs using POST requests in the API (Application Programming Interface), i.e. delivering information (to a server) in a structured format in order to receive a response in return. In the case of GAI, the response is generated by the called-upon LLM model. In the case of GPT-4 and related models, this process is also known as ``ChatCompletion".

Following is a pair of the form (questions, image) with retrieved response. The following are taken from Quiz 1, General Pathology - Atherosclerosis \& Thrombosis (Question 1) and from Quiz 3, Organ System Pathology - Dermatopathology  (Question 2):
\begin{itemize}
    \item Question 1: A 65-year-old man has had increasing dyspnea and orthopnea for the past year. On physical examination there are rales in all lung fields. A chest x-ray shows cardiomegaly and pulmonary edema. An echocardiogram shows reduced cardiac output. The gross appearance of his heart shown here is most consistent with which of the following underlying conditions?
    \begin{itemize}
    \begin{figure}[ht]
    \centering
            \includegraphics[width=0.20\textwidth]{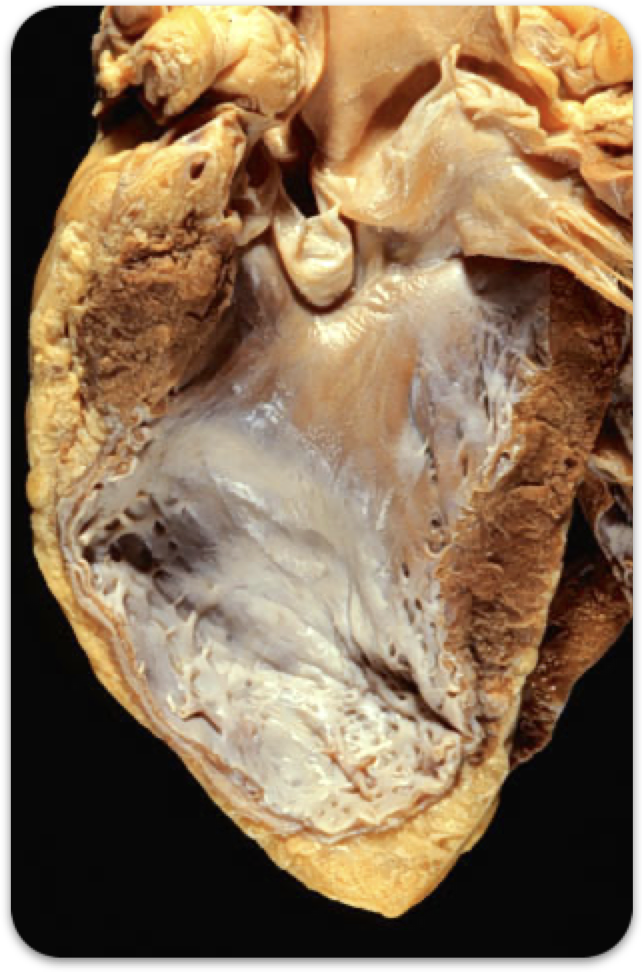}
            \caption{Associated Image to symptoms described in Question 1(CV)}
            \label{fig:cvq1}
        \end{figure}
        \item A. Amyloidosis
        \item B. Systemic hypertension
        \item C. Diffuse scleroderma
        \item D. Atherosclerosis
        \item E. Viral myocarditis  
    \end{itemize}
    \item Question 2: A 36-year-old man incurs a traumatic laceration to his right upper chest. This is repaired. Over the next 3 months, he notes the development of the lesion shown here, which was excised. Microscopic examination of this lesion is most likely to show which of the following findings?
    \begin{itemize}
    \begin{figure}[ht]
    \centering
            \includegraphics[width=0.35\textwidth]{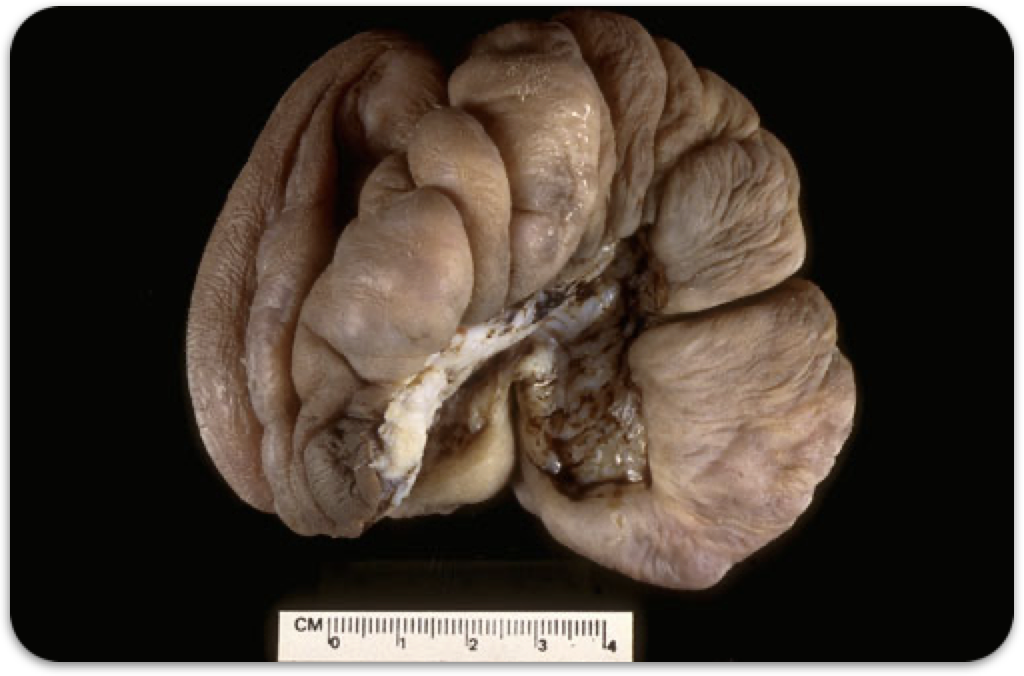}
            \caption{Associated Image to symptoms described in Question 2(SKIN)}
            \label{fig:Dermq2}
        \end{figure}
        \item A. Necrotizing acute inflammation
        \item B. Granulomas with caseous necrosis
        \item C. Apoptosis
        \item D. Dense collagen bundles
        \item E. Atypical squamous epithelium 
    \end{itemize}   
\end{itemize}

The Configuration and Rules of Conduct used through out the interactions with GPT-4-Vision-Preview (which was the LLM selected for multimodal evaluation) are the following: 

\begin{lstlisting}[language=Python]
MAX\_TOKENS = ``4000''
image\_folder = \text{pkg\_resources.resource\_filename (``{visionGptEvaluation'', ``static/images/'')
rule_of_conduct = "Describe the image and then use the that description and the following symptoms to choose the correct answer without explanation. At the end of the response write Correct Choice:(ONLY the correct letter)."
API\_URL = ``{https://api.openai.com/v1/chat/completions}''
MODEL = ``gpt-4-vision-preview''
\end{lstlisting}

The correct responses are evaluated based on the response returned by the selected LLM, while the incorrect responses are evaluated based on the default explanations provided by the domain experts. In the case of this study, explanations are provided for the responses by the creators of the MCQs and are available online \cite{utah}.

It is important to note that transformer architectures, like the one used in LLMs, incorporate both question-answer dynamics and, in the case of multimodal applications, image analysis capabilities. In such multimodal systems, the model processes and integrates information from different sources. Specifically, it considers the textual input, such as a question or prompt, in conjunction with visual data from images. This integrated approach allows the model to generate responses that are informed by a combination of linguistic and visual context, enhancing its ability to understand and interact in more complex scenarios. 

Following are two examples of question structure, image, response and explanation. In the first example, the LLM returned an erroneous answer, while the right answer was returned in the second example.
\begin{itemize}
  \item Example 1 \ref{fig:cvq1}
    \begin{itemize}
    
    \item LLM's response (Question 1): The image shows a gross Pathology specimen of a heart that appears enlarged and has thickened walls, particularly in the left ventricle. There are no signs of nodules, extensive fibrosis, or inflammatory lesions indicative of the other conditions listed. Choice:B
    \item Explanation of solution (Question 1, Correct Choice: D): This is a ventricular aneurysm that resulted from healing of a large myocardial infarction resulting from severe occlusive coronary artery atherosclerosis. 
    \end{itemize}
    \item Example 2  \ref{fig:Dermq2}
    \begin{itemize}
        \item LLM's response (Question 2): The image shows a large, irregular, flesh-colored mass with a rough texture. It appears to be an excised piece of tissue, with some areas that look fibrous and others that look softer and possibly mucoid or fatty. The ruler in the image provides a scale indicating the lesion is several centimeters across. Given the history of a traumatic laceration, the subsequent development of this lesion over months, and the appearance of the tissue, which suggests excessive growth possibly due to scar formation, microscopic examination is most likely to show dense collagen bundles, which are indicative of a scar or keloid formation.Choice:D
        \item Explanation of solution (Question 2, Correct Choice:D):This is a keloid, or hypertrophic scar. This lesion results from an exaggerated healing reaction in which there is marked deposition of collagenous tissue in the dermis. 
    \end{itemize}
\end{itemize}


\subsection{Multimodal evaluation}
\begin{figure}[H]
    \centering
    \includegraphics[width=1.0\linewidth]{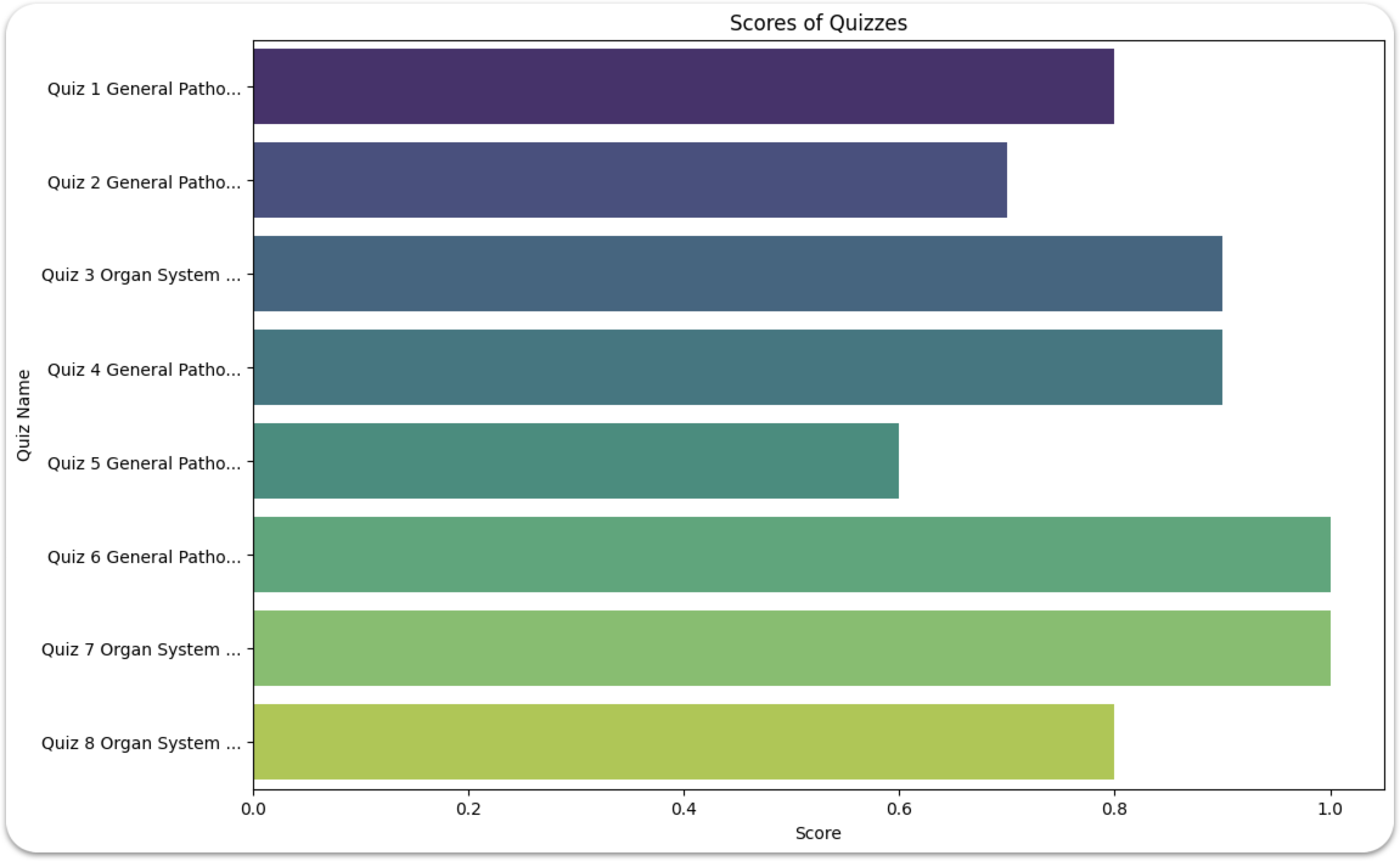}
    \caption{Total Scores per Domain}
    \label{fig:anal}
\end{figure}

In figure~\ref{fig:anal}, we show the results from evaluation of the LLM correctness in the different sub-domains of Pathology. This graph is based on the calculated scores on responses received by GPT-4-Vision-Preview on the corresponding quizzes. In detail:
\begin{enumerate}
 \item General Pathology
 \begin{enumerate}
     \item Atherosclerosis \& Thrombosis: (Quiz 1)
     For Quiz 1 10 multiple-choice questions with images were provided and 8 out of 10 were correctly answered. 
    \item Cell Injury (Quiz 2)
    For Quiz 2 10 multiple-choice questions with images were provided and 7 out of 10 were correctly answered. 
    \item ImmunoPathology (Quiz 4)
    For Quiz 4 10 multiple-choice questions with images were provided and 9 out of 10 were correctly answered.
    \item Inflamation (Quiz 5)
    For Quiz 5 10 multiple-choice questions with images were provided and 6 out of 10 were correctly answered.
    \item Neoplasia (Quiz 6 - 10 questions)
    For Quiz 6 10 multiple-choice questions with images were provided and 10 out of 10 were correctly answered.
 \end{enumerate}
   \item Organ System Pathology 
   \begin{enumerate}
     \item Cardiovascular Pathology (Quiz 7)
     For Quiz 7 9 multiple-choice questions with images were provided and 9 out of 9 were correctly answered.
    \item DermatoPathology (Quiz 3)
    For Quiz 3 10 multiple-choice questions with images were provided and 9 out 10 were correctly answered. 
    \item Endocrine Pathology (Quiz 8 - 10 questions)
    For Quiz 7 10 multiple-choice questions with images were provided and 8 out of 10 were correctly answered.
 \end{enumerate}
\end{enumerate}
The total score retrieved from the provided quizzes amounted to approximately 84\%. In 79 questions paired with images, 66 were correctly answered, while for 13 an erroneous answer was provided.

\subsection{Domain-specific analysis of results}
\subsubsection{Image Metadata Analysis}
The images featured in the MCQ depict various organs, utilizing diverse imaging techniques and technologies. To accurately determine the strengths and weaknesses of the LLM being assessed, we categorize each image based on its general domain, with relevant abbreviations included as part of the file's metadata. 

\begin{figure}[H]
    \centering
    \includegraphics[width=0.9\linewidth]{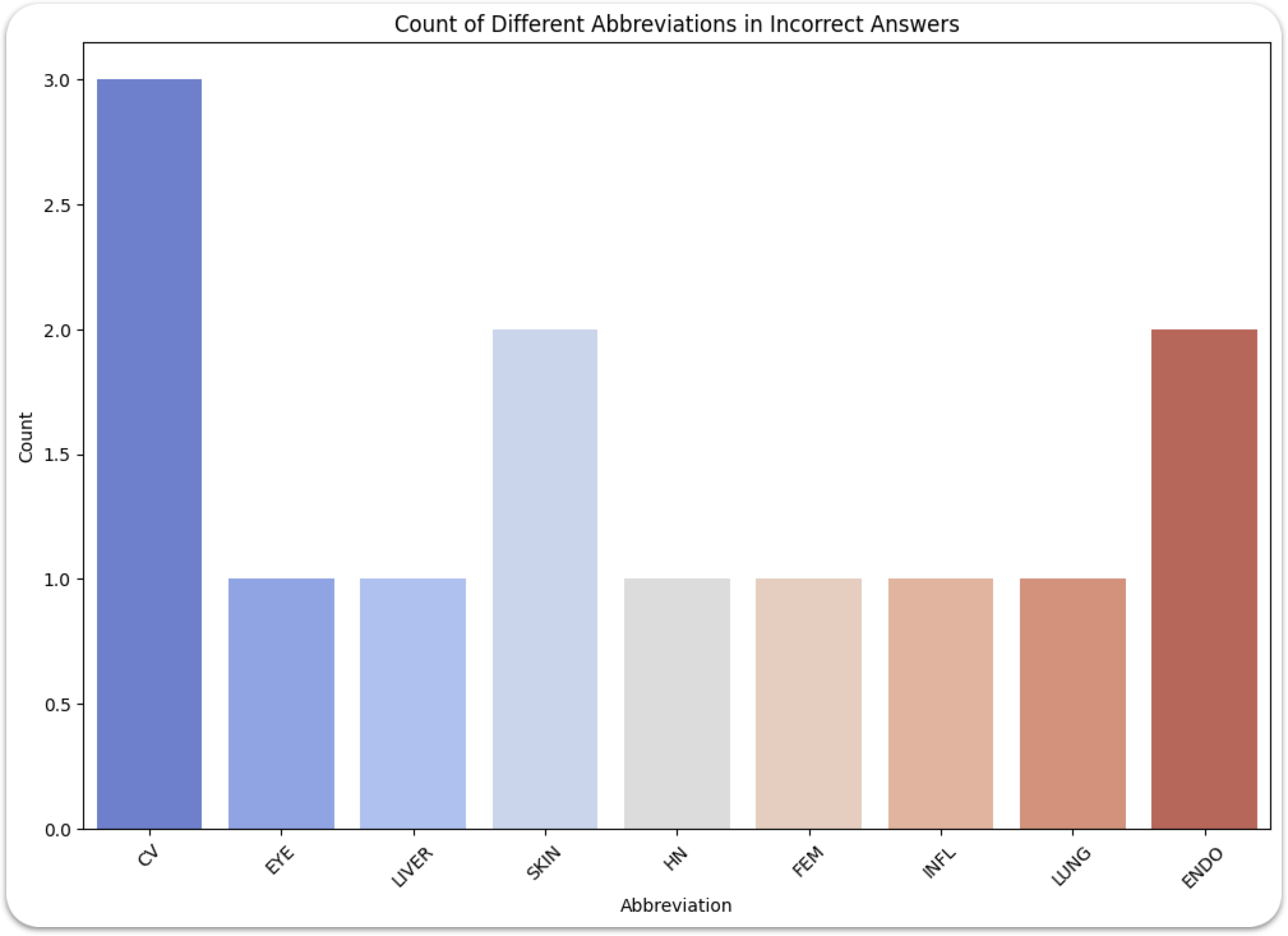}
    \caption{Incorrect responses based on images used}
    \label{fig:incab}
\end{figure}

\begin{figure}[H]
    \centering
    \includegraphics[width=0.9\linewidth]{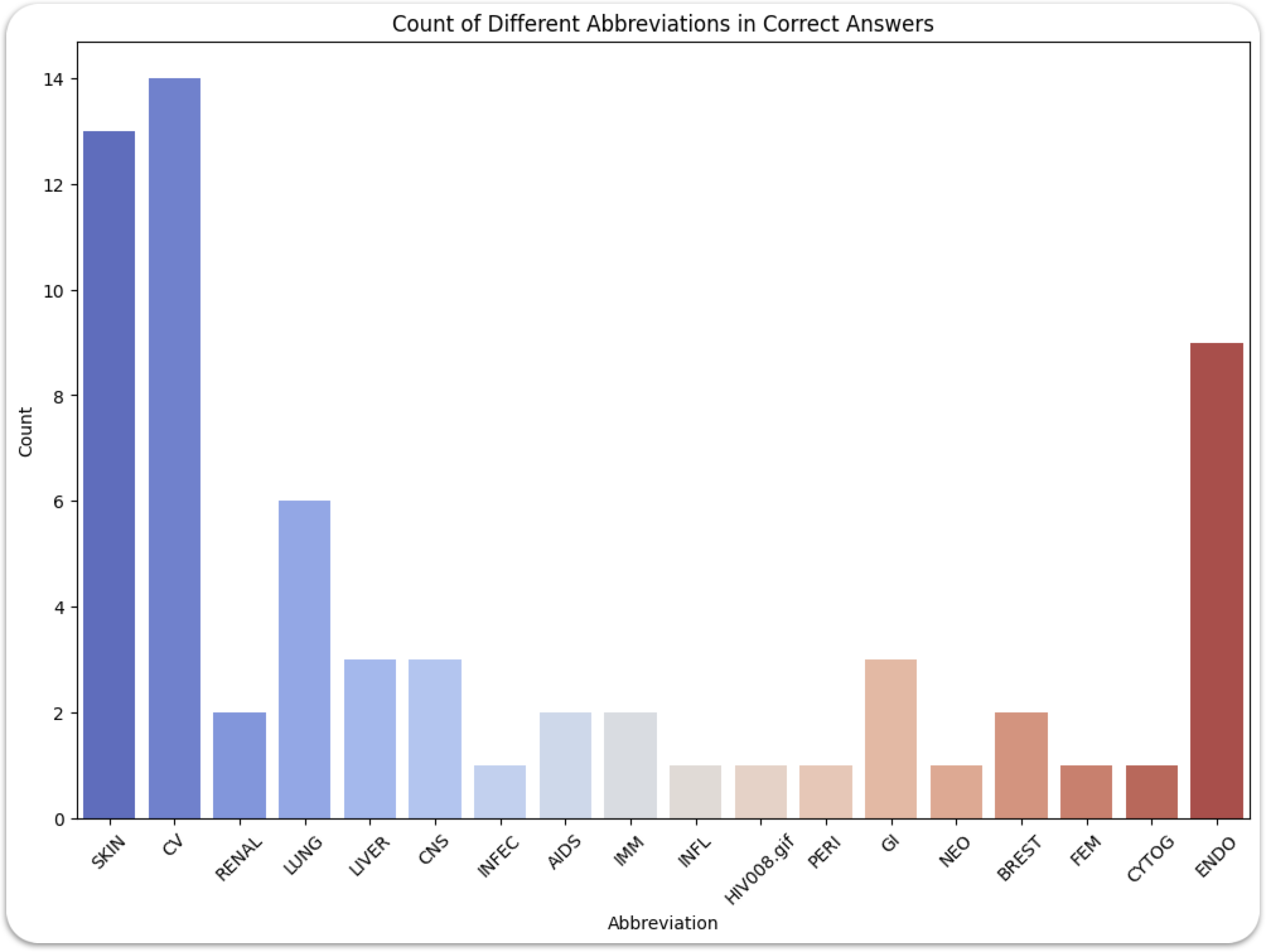}
    \caption{Correct responses based on images used}
    \label{fig:corab}
\end{figure}

In more detail, for the incorrect responses-choices provided by the LLM, as can be seen in figure \ref{fig:incab}, the images were related to:
\begin{enumerate}
    \item CV (3 images): CV stands for ``Cardiovascular" referring to the heart and blood vessels. Cardiovascular imaging involves various techniques and procedures to visualize the heart and blood vessels, assess their structure and function, and diagnose cardiovascular diseases. This can include modalities like echocardiography, angiography, computed tomography (CT), magnetic resonance imaging (MRI), and nuclear medicine imaging techniques. Each of these methods provides different insights into the cardiovascular system's health and function, aiding in the diagnosis and management of various heart and vascular diseases.
    \item Eye (1 image): When it comes to the ``eye" several specialized techniques are used to visualize, diagnose, and monitor conditions affecting the eyes. The eye is a complex and delicate organ responsible for vision, and accurate imaging is essential for diagnosing and treating ocular conditions.
     \item Liver (1 images): In medical imaging, when referring to the ``liver," it involves various techniques used to visualize, diagnose, and monitor liver conditions and diseases. The liver is a vital organ that performs essential functions, including detoxification, protein synthesis, and production of biochemicals necessary for digestion.
    \item Skin (2 images): Various techniques are employed for diagnosing, monitoring, and treating skin conditions. Imaging includes among others Dermoscopy, Optical Coherence Tomography, Confocal Microscopy, Ultrasound Imaging, Photography and Thermography.
    \item HN (1 image): HN refers to head and neck, and it is more commonly used in radiology. It usually refers to imaging related to MRI, CT scans and X-rays, with a focus on head and reck region of the body.
    \item FEM (1 image): Referring to Femur or Femoral, the thigh bone or the surrounding region. 
    \item INFL (1 image): Images that provide visual evidence of inflammation in various body tissues and organs. 
    \item LUNG (1 image): Images related to the lungs.
    \item ENDO (2 image): Endocrine medical imaging focuses on diagnosing and evaluating diseases and conditions affecting the endocrine system. The endocrine system includes a collection of glands that produce hormones, which regulate many of the body's functions.
\end{enumerate}
In figure~\ref{fig:corab}, the images are highlighted along with related abbreviations used to provide the correct responses. Most of the used images are related to Skin (13 images), or the Cardiovascular (14 images) and the Endocrine system (9 images). There is a good distribution of images related to other organs. The totality of the corresponding abbreviations are included in the abbreviation list.

It should also be noted that images related to the Eye and the Head and Neck were only present in incorrect responses. 

\subsubsection{Named Entity Recognition Analysis}
In this section, the NER part of the analysis is described in detail. Since there was a vast amount of extracted information, only the most prominent parts of the analysis are highlighted and more can be provided upon request.

The entities are organized into `Entity Types' and `Entity Names' as shown in Table~\ref{tab:entIt}. An `Entity Type' can be a disease, a condition, an organ, or another category related to the domain. The `Entity Name' refers to the specific object identified within its corresponding type. Both are grouped, depending of whether they belong to a correct or an incorrect response, by response or explanation respectively, as already described in the previous section and shown in figure~\ref{fig:analProc}.

\begin{table}[H]
    \centering
    \begin{tabular}{ccc}
         ENTITY TYPE& ENTITY NAME & GROUP\\
        DISEASE & Myocardial Infarction  & 0 \\
        CONDITION&	Severe Occlusive&	0 \\
         .. & .. & .. \\
         DISEASE &	Atherosclerosis &	1 \\
BODY PART	&Lower Abdominal Aortic &	1 \\
CONDITION	&Atherosclerotic Aneurysm &	1 \\
    \end{tabular}
    \caption{Snippet of medical Entities used for NER}
    \label{tab:entIt}
\end{table}

\begin{figure}[h]
    \centering
    \includegraphics[width=1.0\linewidth]{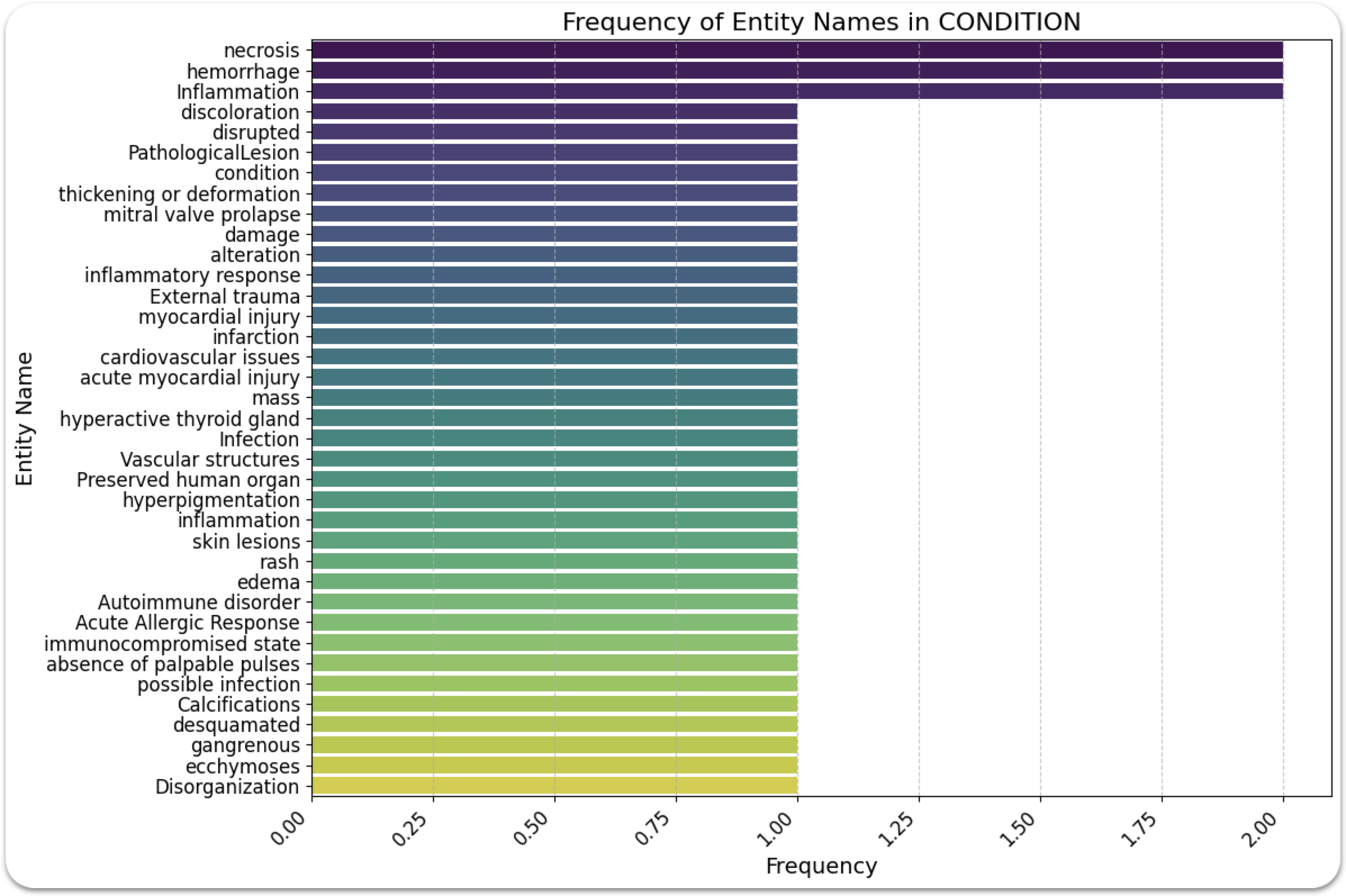}
    \caption{Frequency of identified Conditions in Correct Responses}
    \label{fig:condCor}
\end{figure}

\begin{figure}[h]
    \centering
    \includegraphics[width=1.0\linewidth]{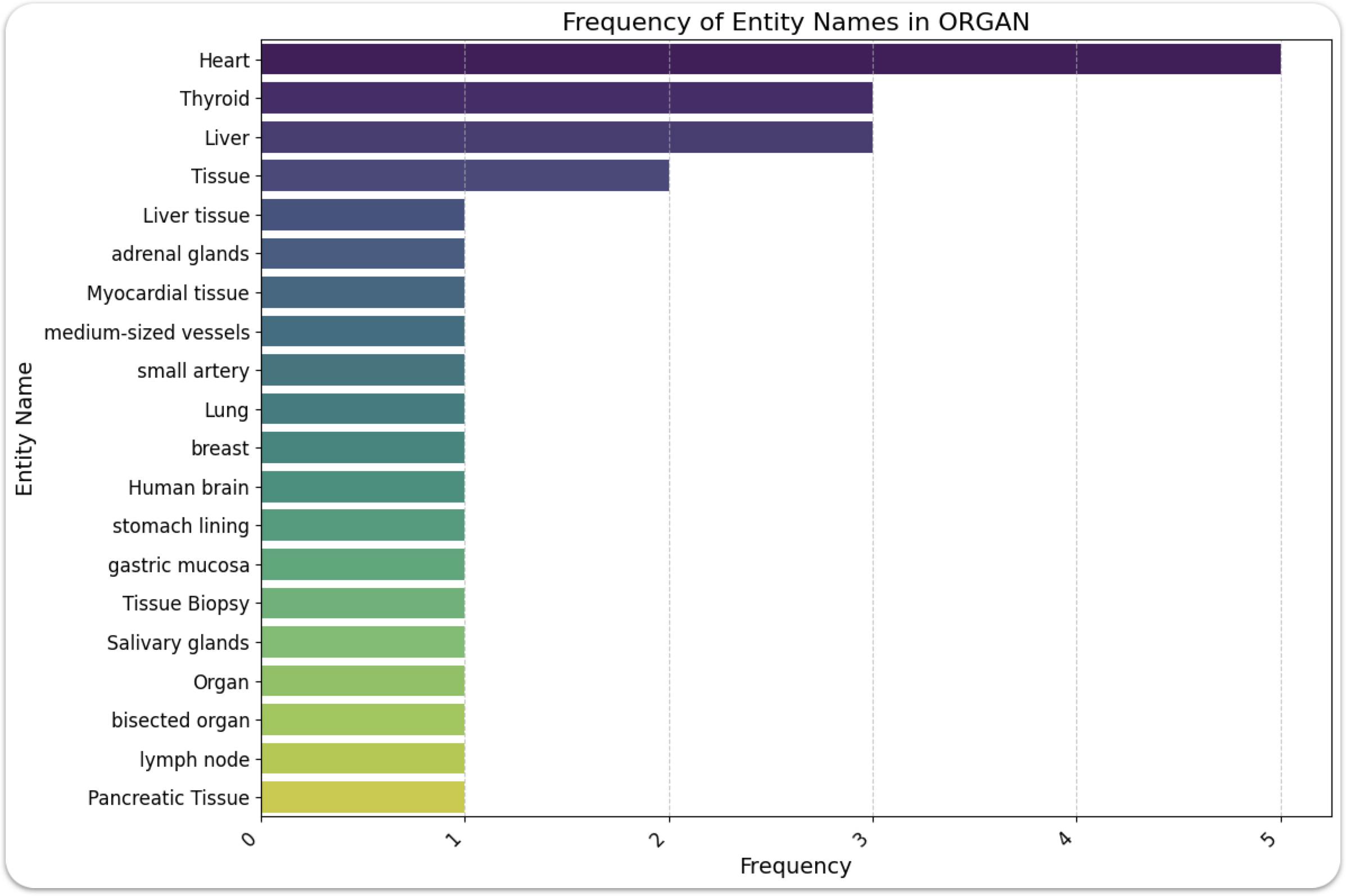}
    \caption{Frequency of identified Organs in Correct Responses}
    \label{fig:orgCor}
\end{figure}

The data, as illustrated in Figures~\ref{fig:condCor} and~\ref{fig:orgCor}, prominently showcase the broad range of the encompassed domains. We also emphasize the domains related to organs and conditions in the solution explanations for the incorrect responses. 

\begin{figure}[h]
    \centering
    \includegraphics[width=1.0\linewidth]{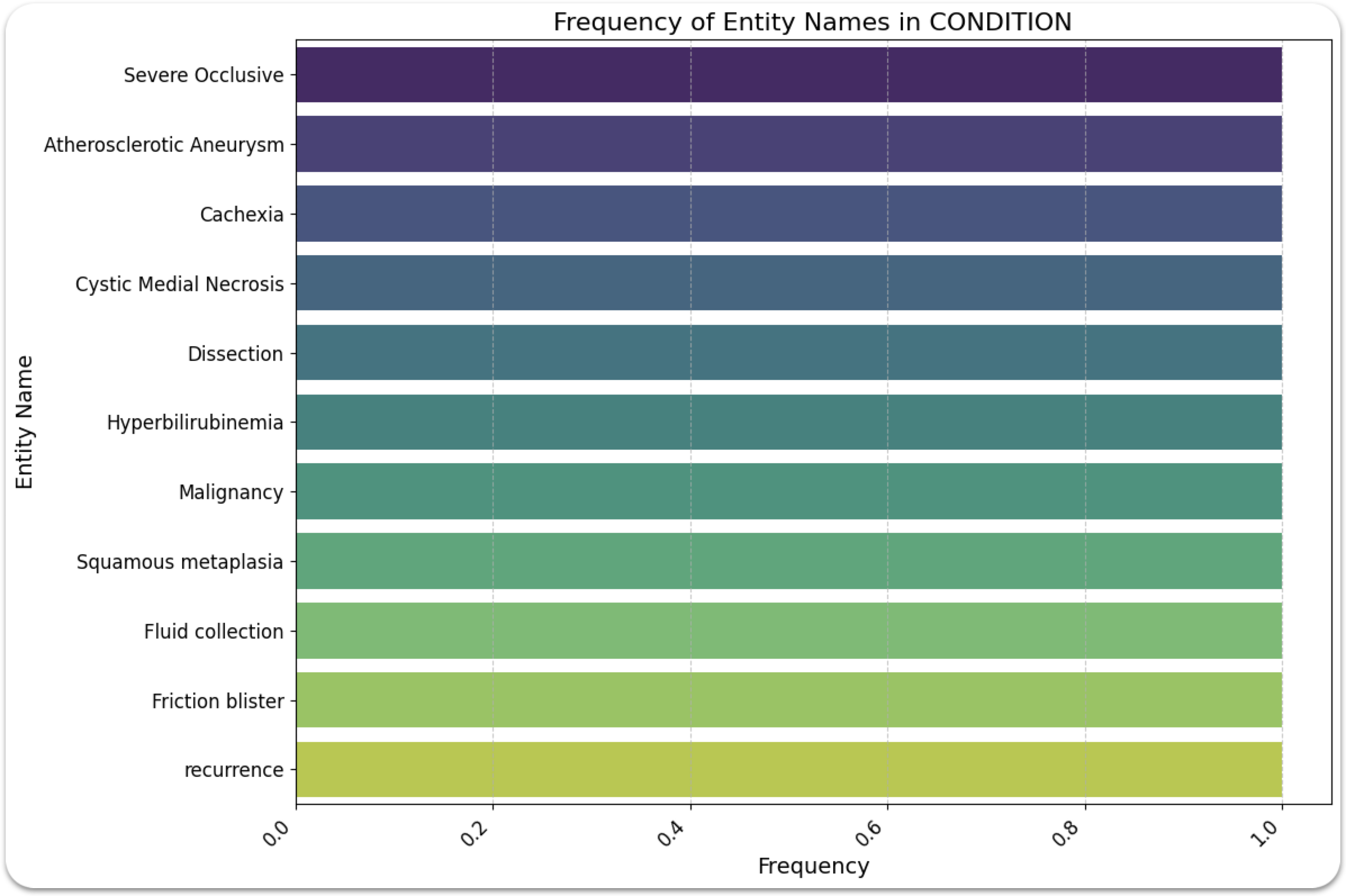}
    \caption{Frequency of identified Conditions in Explanation of solution related to Incorrect Responses}
    \label{fig:condInc}
\end{figure}

As can be seen in figure~\ref{fig:condInc}, the conditions present in the incorrect responses, extracted by the solution explanations, are related to Severe Occlusive, Atherosclerotic Aneurysm, Cachexia, Cystic Medial Necrosis, Dissection, Hyperbilirubinemia, Malignancy, Squamous metaplasia, Fluid Collection, Friction Blister and Recurrence. These conditions, are interconnected primarily through their underlying patho-physiological processes or are manifestations or complications of underlying diseases \citep{hanke2001discovery}. Some interconnections are listed below:

\begin{itemize}
    \item Patho-physiological Processes: Atherosclerotic aneurysms involve the weakening of blood vessel walls, while cystic medial necrosis affects the elasticity of the aorta. Dissection is a consequence of such weakening, leading to a split in the vessel wall.
    \item Manifestations of Underlying Diseases: Cachexia often occurs in the advanced stages of malignancy. Similarly, hyperbilirubinemia might be a symptom of liver diseases, which could also be associated with malignancies.
    \item Complications and Consequences: Some conditions listed are complications or consequences of other diseases. Fluid collection can occur as a result of inflammation, infection, or injury. A friction blister is a direct response to skin trauma.
    \item Chronic and Acute Conditions: The list includes both chronic conditions (like malignancy) and acute conditions (like dissection). This highlights the spectrum of medical conditions ranging from long-term diseases to immediate medical emergencies.
    \item Recurrence: The term `recurrence' is a common theme in many diseases, where the condition reappears after a period of remission or apparent cure. This is often a concern in malignancies, but it can apply to other conditions as well.
\end{itemize}

In summary, while these conditions are distinct in their specific characteristics and the body systems they affect, they are interrelated through their underlying patho-physiological mechanisms, their role as indicators of other health issues, their nature as either chronic or acute conditions, and the possibility of recurrence.

\begin{figure}[H]
        \centering
        \includegraphics[width=1.0\linewidth]{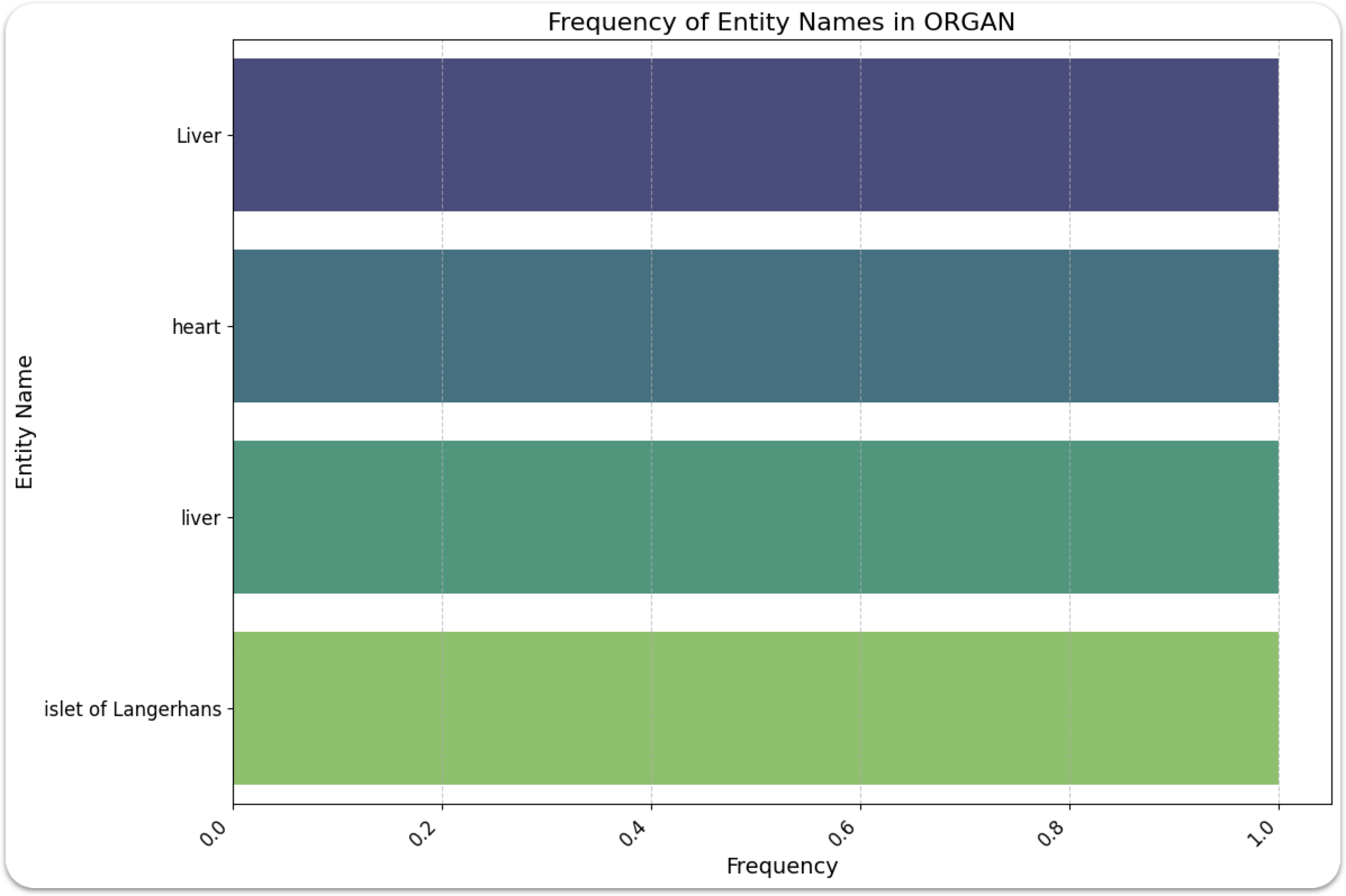}
        \caption{Frequency of identified Organs in Explanation of solution related to Incorrect Responses}
        \label{fig:orgInc}
    \end{figure}

\subsubsection{Knowledge Graphs}
The NER analysis is followed by the construction of KGs based on the returned responses for the correct answers and the solution explanations for the case of the incorrect answers.

\begin{figure}[H]
    \centering
    \includegraphics[width=1\linewidth]{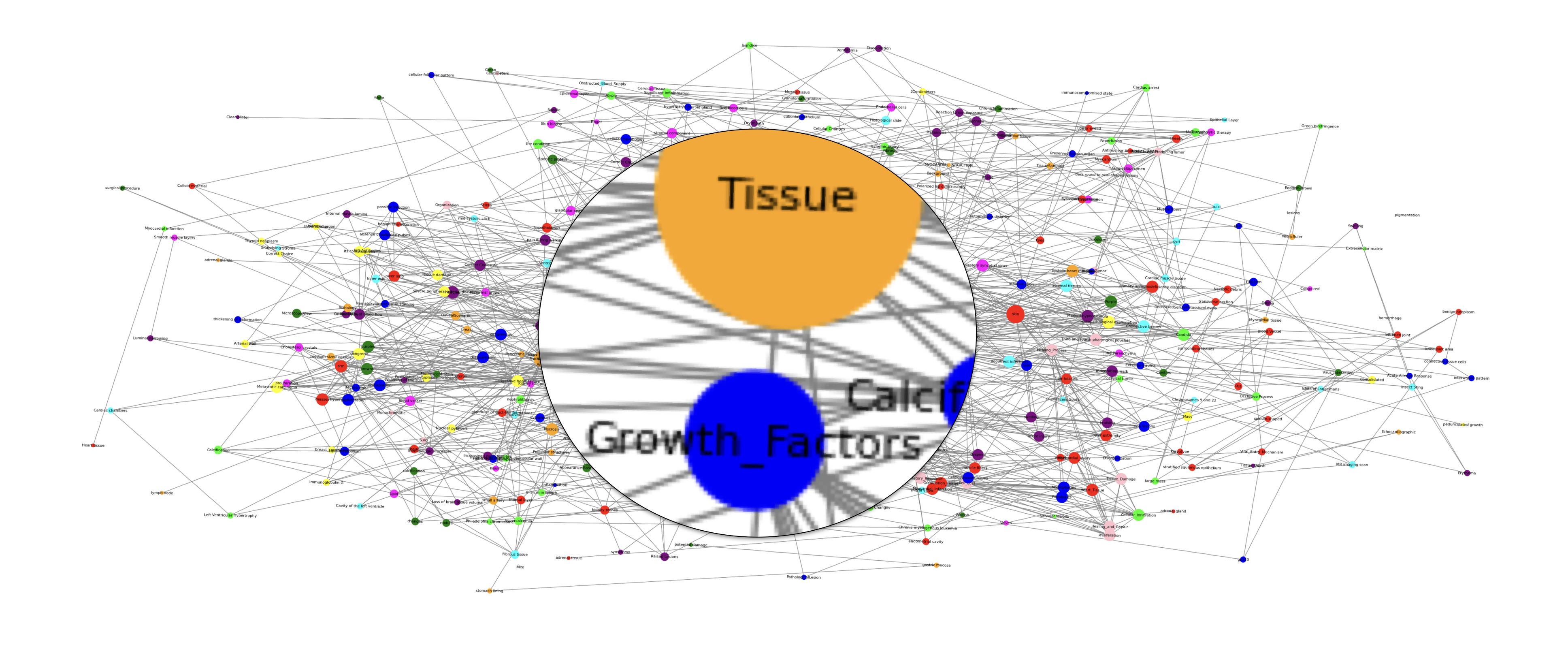}
    \caption{Knowledge Graph of Returned Responses}
    \label{fig:corMp}
\end{figure}
With regard to the correct responses, the main quantitative measures of the KG in figure~\ref{fig:corMp} are the following:
\begin{itemize}
    \item Network Density: 0.01675
    
    The density of a network is the proportion of possible edges that are actual edges in the network. A low density, such as the calculated 0.0163, suggests that the network is quite sparse, meaning that the test-taker made relatively few associations between the concepts they identified correctly.
    
    \item Number of Connected Components: 25
    
     A connected component is a set of nodes in which each pair of nodes are connected via a path. Having 25 connected components suggests that the LLM's correct answers are distributed across various unconnected groups of concepts, indicating a segmented understanding of the material.
     
     \item Top Nodes by Degree:   
     The degree of a node is the number of connections it has to other nodes. `Tissue' with the highest degree (45) suggests it was the concept most frequently identified or associated with other concepts correctly. The other top nodes like `skin', `cells', `nuclei', and `edema' with high degrees suggest these are also central to the test-taker's correct answers.
     \begin{itemize}
         \item Tissue: 45
         \item Skin: 29
         \item Cells: 27
         \item Nuclei: 23
         \item Edema: 22
     \end{itemize}   
\end{itemize}

\begin{figure}[H]
    \centering    \includegraphics[width=1\linewidth]{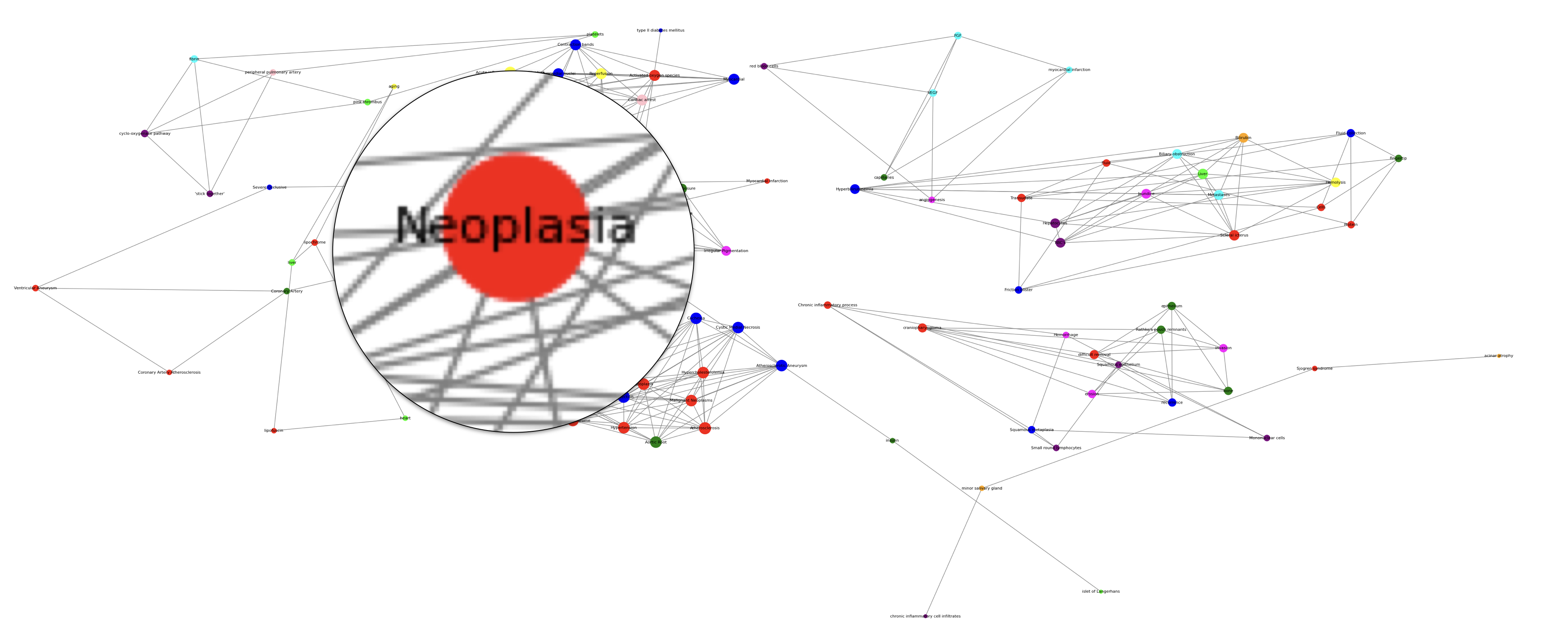}
    \caption{Knowledge Graph of Solution Explanation for the incorrect answers}
    \label{fig:incorMp}
\end{figure}

With regard to the incorrect responses, the main quantitative measures of the KG in figure~\ref{fig:incorMp} are the following: 

\begin{itemize}
    \item Network Density: 0.0633
    
    A higher network density, such as the calculated 0.0633, indicates that the test-taker made more associations between the concepts they got wrong compared to the ones they got right. Thus, the incorrect responses are more associated and can be addressed more easily.
    
    \item Number of Connected Components: 13
    
     The lower number of connected components (13) in the solution explanations show a more interconnected understanding of the material, thus revealing greater relation in the material on which the LLM failed.
     
     \item Top Nodes by Degree: 
     The nodes with the highest degree in the incorrect answers relate to cardiovascular Pathology (Atherosclerosis, Dissection, Lower abdominal aortic, Atherosclerotic aneurysm) and Diabetes mellitus. These are areas where the LLM exhibits particular misunderstandings or difficulties.
     \begin{itemize}
         \item Atherosclerosis: 12
         \item Dissection: 12
         \item Lower abdominal aortic: 11
         \item Atherosclerotic aneurysm: 11
         \item Diabetes mellitus: 11
     \end{itemize}   
\end{itemize}

The explanation in the incorrect responses - knowledge graph, indeed reveal patterns of misconceptions and could guide targeted interventions. 

\section{Discussion}

In our recent prior research, we extensively evaluated GPT-4’s capabilities in symptom analysis, either (1) under medical expert supervision or (2) through sophisticated MCQs, used for education purposes to evaluate medical students. With regard to the MCQs based on symptomatology in the domain of General Pathology, GPT-4 achieved over 90\% accuracy in approximately 180 questions.
\subsection{Summary of results}
In this work, we examined GPT-4-Vision-Preview capabilities in image analysis paired with symptom analysis in different domains of Pathology. Our evaluation again used MCQs. The total score retrieved was approximately 84\%. In 79 questions paired with images, 66 were correctly answered, while 13 were erroneously answered.

The main components and highlights of our research are the following:

\begin{enumerate}
    \item Methodology
\begin{itemize}
    \item Multimodal LLM Evaluation: Overall, this approach aims at enhancing the evaluation of LLMs with vision by using structured tests that combine images and related questions. This challenges the LLM to interpret visual (static image) content and textual information simultaneously, determining the most appropriate answer from the given choices. We provide the necessary steps for data pre-processing, interaction engineering and data extraction. 
    \item  Domain-Specific Analysis: This consists of three steps. The first step involves IMA, where the model's responses are evaluated to determine the domains in which it responds correctly or incorrectly. The second step applies NER to both the correct and incorrect responses. For the correct responses, this involves analyzing the response itself. For the incorrect responses, it involves examining the solution explanation. This step generates a report that categorizes entities based on their association with either a response or an explanation. The third and final step involves plotting the extracted entities onto KGs. This step's report focuses on assessing the density of connections and the potential for information spread within these graphs.
\end{itemize}
    \item Evaluation of Multimodal LLM Competency in the Medical Domain: Using Pathology MCQs, we found that GPT-4-Vision-Preview has great understanding in questions and images related to Pathology, scoring 84\% in the provided questions.
    \item Knowledge Extraction to drive Optimisation: Using the methodology described previously, we investigated the results and found important paths that could lead to fast and accurate interventions and, thus, lead to improved outcomes. Namely:
    \begin{enumerate}
        \item According to the IMA most inconsistencies were related to images in the CardioVascular domain (CV - 3 images), Skin (SKIN - 2 images) and the Endocrine system (ENDO - 2 images).
        \item According to NER analysis, inconsistencies were found in distinct conditions that are interconnected through their underlying patho-physiological mechanisms, roles as health indicators, and chronic or acute nature, including recurrence potential.
        \item Based on the KGs, we found areas where the LLM exhibits particular misunderstandings or difficulties, especially in relation to related to cardiovascular Pathology (Atherosclerosis, Dissection, Lower abdominal aortic, Atherosclerotic aneurysm) and Diabetes Mellitus.
    \end{enumerate}

\end{enumerate}
These findings are important to drive targeted interventions for LLM optimisation in the examined domain. Furthermore the proposed methodology can also be generalised and used in other domains and, thus, drive tailored and targeted fine-tuning approaches.
\subsection{Future work}

The assessment and training of medical students is a multifaceted endeavor, underscoring the fact that the MCQs employed in this research constitute merely a fraction of the comprehensive evaluation required for doctor's certification. The notable precision of a multimodal LLM in accurately identifying and categorizing various diseases, conditions, and health states, based on symptom descriptions and pertinent images drawn from the extensive scope of human pathology, highlights the direction of our future research. Thus, in our future work we aim at exploring additional methods to simulate and assess multimodal LLMs' effectiveness in a broader range of evaluative processes of the medical domain.

\section*{Acknowledgements}
This work has been partly supported by the University of Piraeus Research Center.

The authors would also like to extend their gratitude to the doctors of Dermacen SA~ {\url{www.dermatologikokentro.gr/}} for their technical assistance and guidance as per the various medical themes reported in this work. 

\section*{Data Availability}
Data sources utilized in this study have been cited and referenced. We are committed to contributing to the scientific community by open-sourcing both the engineered data and the code developed for this research, which will be made publicly available concurrent with the publication of our findings.

\section*{Abbreviations}
The following abbreviations are used in this paper:\\
\begin{tabular}{@{}ll}
ML & Machine learning \\
AI & Artificial Intelligence \\
IMA & Image Metadata Analysis \\
KG & Knowledge Graphs \\
XAI & Explainable Artificial Intelligence \\
NLP & Natural Language Processing\\
LLM & Large Language Model \\
MCQ & Multiple-choice Questionnaire \\
NER & Named Entity Recognition\\
ViT & Vision Transformer\\
CNN & Convolutional Neural Network\\
SKIN & Refers to conditions or diseases related to the skin\\
CV & Cardiovascular, relating to the heart and blood vessels\\
RENAL & Relating to the kidneys\\
LUNG & Pertaining to the lungs\\
LIVER & Relating to the liver\\
CNS & Central Nervous System \\
INFEC & Infection \\
AIDS - HIV & Acquired Immune Deficiency Syndrome, a disease caused by the HIV virus \\
IMM & Immunology\\
INFL & Inflammation\\
PERI & Perinatal \\
GI & Gastrointestinal\\
NEO & Neoplastic \\
BREAST & Chest and breast \\
FEM & Femoral, relating to the thigh or femur bone\\
CYTOG & Cytogenetics\\
ENDO & Endocrinoloy or endocrine system \\
\end{tabular}

\bibliography{ref2}

\end{document}